\documentclass{midl} 

\usepackage{mwe} 
\usepackage{graphicx,verbatim}
\usepackage{graphicx}
\usepackage{color}
\usepackage{url}
\usepackage{booktabs}
\usepackage{amsmath}
\usepackage{xcolor,colortbl}
\definecolor{lavender}{gray}{0.9}
\usepackage{soul}
\usepackage{bbding}
\usepackage{graphicx,verbatim}
\usepackage{amssymb}
\usepackage{bbm}

\usepackage[dvipsnames]{xcolor}
\definecolor{lavender}{gray}{0.9}
\definecolor{supervised}{RGB}{245,245,245}      
\definecolor{semisup}{RGB}{236,244,252}        
\definecolor{crosssup}{RGB}{243,238,252}       
\definecolor{upper}{RGB}{235,242,235}          

\usepackage{amsmath} 
\jmlrvolume{-- Under Review}
\jmlryear{2026}
\jmlrworkshop{Full Paper -- MIDL 2026 submission}
\editors{Under Review for MIDL 2026}

\title[Endo-SemiS]{Endo-SemiS: Towards Robust Semi-Supervised Image Segmentation for Endoscopic Video}

\midlauthor{\Name{Hao Li}\midljointauthortext{Corresponding author}\nametag{$^{1}$}\Email{hao.li.1@vanderbilt.edu}\\
\Name{Daiwei Lu\nametag{$^{1}$}} \Email{daiwei.lu@vanderbilt.edu}\\
\Name{Xing Yao\nametag{$^{1}$}} \Email{xing.yao@vanderbilt.edu}\\
\Name{Nicholas Kavoussi\nametag{$^{2}$}} \Email{nicholas.l.kavoussi@vumc.org}\\
\Name{Ipek Oguz\nametag{$^{1}$}} \Email{ipek.oguz@vanderbilt.edu}\\
\addr $^{1}$ Vanderbilt University \addr \\$^{2}$ Vanderbilt University Medical Center
}

\begin{document}

\maketitle

\begin{abstract}
In this paper, we present \textbf{Endo-SemiS}, a semi-supervised segmentation framework for providing reliable segmentation of endoscopic video frames with limited annotation. Endo-SemiS uses 4 strategies to improve performance by effectively utilizing all available data, particularly unlabeled data: (1) Cross-supervision between two individual networks that supervise each other; (2) Uncertainty-guided pseudo-labels from unlabeled data, which are generated by selecting high-confidence regions to improve their quality; (3) Joint pseudo-label supervision, which aggregates reliable pixels from the pseudo-labels of both networks to provide accurate supervision for unlabeled data; and (4) Mutual learning, where both networks learn from each other at the feature and image levels, reducing variance and guiding them toward a consistent solution.
Additionally, a separate corrective network that utilizes spatiotemporal information from endoscopy video to improve segmentation performance. Endo-SemiS is evaluated on two clinical applications: kidney stone laser lithotomy from ureteroscopy and polyp screening from colonoscopy. Compared to state-of-the-art segmentation methods, Endo-SemiS substantially achieves superior results on both datasets with limited labeled data. The code is publicly available at \url{https://github.com/MedICL-VU/Endo-SemiS}


\end{abstract}

\begin{keywords}
Comprehensive supervision, uncertainty-guided pseudo-label, spatiotemporal
\end{keywords}

\section{Introduction}
\begin{figure}[t]
\centering
\includegraphics[width=\linewidth]{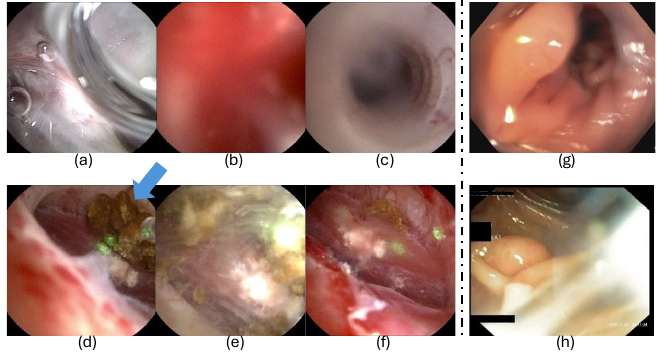}
\caption{Challenging ureteroscopy (a–f, left) and colonoscopy (g–h, right) images for segmentation.
(a) irrigation; (b) bleeding; (c) motion blur;
(d) early ablation; (e) mid ablation;
(f) late ablation. The arrow indicates the target kidney stone for ablation. (g) and (h) are from the public dataset \cite{ali2023multi}, which is collected from multiple imaging sites.}

\label{dataset}
\end{figure}

Endoscopic image segmentation poses unique challenges, including large variations in image quality and appearance, which may be caused by motion blur, fluctuating lighting conditions \cite{li2025automated}, and often fluid-filled environments \cite{setia2023computer}, as well as domain shifts \cite{ali2023multi}. These effects are illustrated in Fig.~\ref{dataset}, which shows blur, bleeding, debris, occlusions, and cross-site or cross-device appearance changes in ureteroscopy and colonoscopy images. The limited availability of manual labels further complicates the task.

Semi-supervised learning (SSL) approaches provide a potential solution by effectively leveraging information from unlabeled data \cite{sohn2020fixmatch,chen2021semi,luo2022semi-crossteaching,luo2022semi-uncertainty,yang2023revisiting,tarvainen2017mean,wang2024allspark}. These methods construct supervision signals for unlabeled samples from the predictions of the model itself. A key approach to achieving this is enforcing consistency constraints \cite{tarvainen2017mean}, either through uncertainty-guided self-regularization \cite{sohn2020fixmatch,yang2023revisiting,luo2022semi-uncertainty,wang2024allspark,tarvainen2017mean} or cross-supervision \cite{chen2021semi,luo2022semi-crossteaching} to improve the quality and reliability of pseudo-labels.

Based on these principles, SSL can be broadly categorized into single-network and dual-network frameworks. Single-network approaches enforce consistency under perturbations and regularize pseudo-labels based on uncertainty. \cite{sohn2020fixmatch,yang2023revisiting,wang2024allspark}. However, single model-based method tends to persist in its incorrect predictions, leading to error accumulation. Dual-network approaches maintain two networks that exchange pseudo-labels
for cross-supervision \cite{chen2021semi,luo2022semi-crossteaching} to mitigate confirmation bias \cite{arazo2020pseudo}. Building on this, numerous studies in medical imaging have achieved excellent segmentation performance  \cite{luo2022semi-crossteaching,luo2022semi-uncertainty,wang2023ssl2,yu2019uncertainty,lei2022semi}.

These existing SSL methods have some limitations: \textbf{(1)} Single-network methods lack model-level consistency, which makes them struggle with high-uncertainty samples. \textbf{(2)} Methods that either use the entire uncertainty map or apply a fixed uncertainty threshold treat many unreliable regions as confident, leading to false positives and overfitting to incorrect pseudo-labels. \textbf{(3)} Cross-supervision methods do not explicitly model uncertainty and struggle to filter out unreliable pseudo-labels. Since each model generates pseudo-labels independently, confirmation bias may occur when both models make similar wrong predictions.

\begin{figure}[t]
\centering
\includegraphics[width=\linewidth]{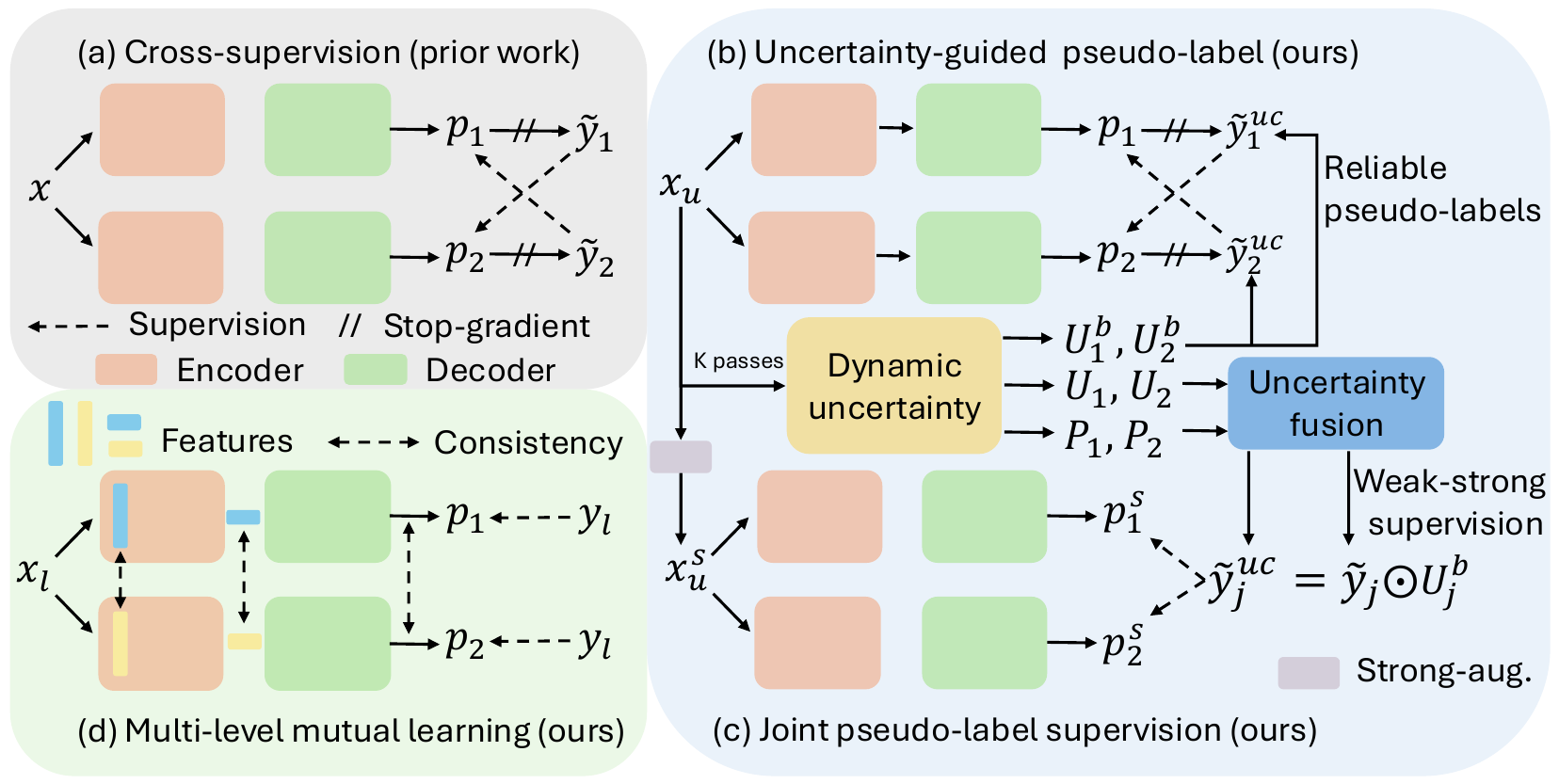}
\caption{The proposed framework adapts the widely used cross-supervision baseline (a) with uncertainty-guided supervision to obtain reliable pseudo-labels (b–c), and further incorporates multi-level mutual learning (d) to improve cross-network consistency. Panels (b–c) (in blue) operate only on unlabeled data $x_u$, whereas (d) is applied only to labeled data $x_l$. The two networks share the same architecture but are optimized independently. $y$, $\tilde{y}$, and $\tilde{y}^{uc}$ denote the ground-truth mask, the raw pseudo-label, and the uncertainty-guided pseudo-label, respectively. $\odot$ denotes the Hadamard (element-wise) product, and $U^b$ is the binary mask from uncertainty map $U$. $x_u^s$ represents a strongly intensity-augmented version of $x_u$.  
We define $\tilde{y}_1^{uc}=\tilde{y}_1\odot U_1^{b}$ and $\tilde{y}_2^{uc}=\tilde{y}_2\odot U_2^{b}$, and omit them for brevity.
}

\label{framework}
\end{figure}

In this paper, we propose \textbf{Endo-SemiS}, a semi-supervised segmentation method to address the limitations of existing approaches in endoscopic imaging with robust outcomes. Specifically, to address each of these limitations: \textbf{(1)} Endo-SemiS adopts a cross-supervision framework (see Fig.~\ref{framework}(a)) to prevent biased learning \cite{chen2022debiased} and uses naive U-Net models to ensure real-time clinical applicability \cite{wei2021shallow,luo2019real} rather than relying on transformer-based models that may require heavy computation \cite{luo2022semi-crossteaching,wang2024allspark}. \textbf{(2)} To obtain reliable pseudo-labels for unlabeled data, a critical step in SSL \cite{wu2021semi}, we leverage both aleatoric and epistemic uncertainty (see Fig.~\ref{framework}(b)). Unlike existing fixed-threshold approaches \cite{sohn2020fixmatch,luo2022semi-uncertainty}, a dynamic thresholding mechanism is applied per uncertainty map, ensuring that only high-confidence regions contribute to pseudo-label supervision. \textbf{(3)} To achieve accurate and consistent supervision, we introduce a joint pseudo-labeling strategy as shown in Fig.~\ref{framework}(c), where supervision is guided by the predictions in the lowest uncertainty regions identified by both networks, and pixels that are classified as uncertain are excluded. \textbf{(4)} We design multi-level mutual learning (see Fig.~\ref{framework}(d)) between networks to further mitigate confirmation bias and improve consistency between networks for producing reliable pseudo-labels. Our main contributions are:

\begin{itemize}
    \item We propose an uncertainty-guided pseudo-labeling approach within a cross-supervision framework, which dynamically filters out unreliable regions for each image and provides more reliable segmentation supervision from unlabeled endoscopic frames.
    \item We introduce a consistency-focused learning framework with joint pseudo-label supervision and multi-level mutual learning. The more reliable prediction between the two networks is selected as supervision, while mutual learning reduces unnecessary prediction variance in confident regions and leads to more stable pseudo-labels.
    \item We design a plug-and-play correction model that uses spatiotemporal information from video to refine segmentation and can be easily integrated into other frameworks.
\end{itemize}

We validate Endo-SemiS on kidney stone laser lithotripsy as a challenging primary task and on polyp screening across different centers to demonstrate generalizability. Our comprehensive evaluation shows consistent improvements over state-of-the-art semi-supervised and fully supervised methods.

\section{Methods}

We begin with a semi-supervised segmentation dataset $D$, which consists of limited labeled data $\{x_l, y_l\}$ and a large amount of unlabeled data $\{x_u\}$, where $x$ and $y$ represent the input images and their annotations, respectively. 
\subsection{Preliminaries}
\paragraph{Generic pseudo-label learning.}
The generic pseudo-label learning \cite{bellver2019budget} for a single network (referred to as Generic) first trains the model $f$, with forward pass $f(\cdot)$ on $\{x_l, y_l\}$ and applies it to $x_u$ to obtain the logit map $f(x_u)$, which is then binarized to form pseudo-label $\tilde{y}_u$ and used as additional supervision. This can be described as: 
\begin{equation}
\label{generic eq}
L = L_s + L_{p}
\end{equation}
where $L_s$ and $L_p$ denote the supervised and pseudo-supervised loss for $\{x_l, y_l\}$ and $\{x_u, \tilde{y}_u\}$.

\paragraph{Cross-supervision.} 
\label{cross-supervision parahraph}
Endo-SemiS employs two individual U-Nets without sharing weights \cite{ronneberger2015u} to achieve cross-supervision signals, as shown in Fig.~\ref{framework}(a). For a given input $x\in\{x_l,x_u\}$, the supervision can be simply extended from Generic (Eq.~\ref{generic eq}) as:
\begin{equation}
\label{pseudo_supervision_eq}
    L_p^{\text{cross}}(x) = L_{p}(f_1(x), \tilde{y}_2) + L_{p}(f_2(x), \tilde{y}_1)
\end{equation}
 where $L_{p}^{\text{cross}}$ represents the cross-supervision applied to both networks using the pseudo-label generated by the other model. The subscripts $i\in\{1,2\}$ indicate the corresponding network. 
 Note that $f_i(x)$ denotes the raw logit map produced by network $i$ for input $x$.
 For brevity, we include it in the loss function term, as it can be converted to probabilities within the loss.

\subsection{Uncertainty-guided pseudo-label}
Uncertainty is introduced into the framework to mitigate confirmation bias (Fig.~\ref{framework}(b)). 
\textit{We hypothesize that uncertainty estimates allow us to identify unreliable pseudo-label regions and exclude them from supervision, so that training focuses on reliable areas.}

\paragraph{Aleatoric uncertainty.}\label{AC} We adopt the widely used weak-to-strong augmentation strategy \cite{sohn2020fixmatch}. Each unlabeled image $x_u$ first undergoes geometric augmentations, referred to as weak augmentation, and $x_u$ is further modified using intensity-based augmentations to obtain a strongly augmented image $x_u^s$. The corresponding pseudo-label $\tilde{y}_u$ is used to supervise the prediction from $x_u^s$. We also leverage CutMix \cite{yun2019cutmix} augmentation on $x_u$ and $x_u^s$ to further increase the robustness and segmentation performance.

\begin{figure}[t]
\centering
\includegraphics[width=\linewidth]{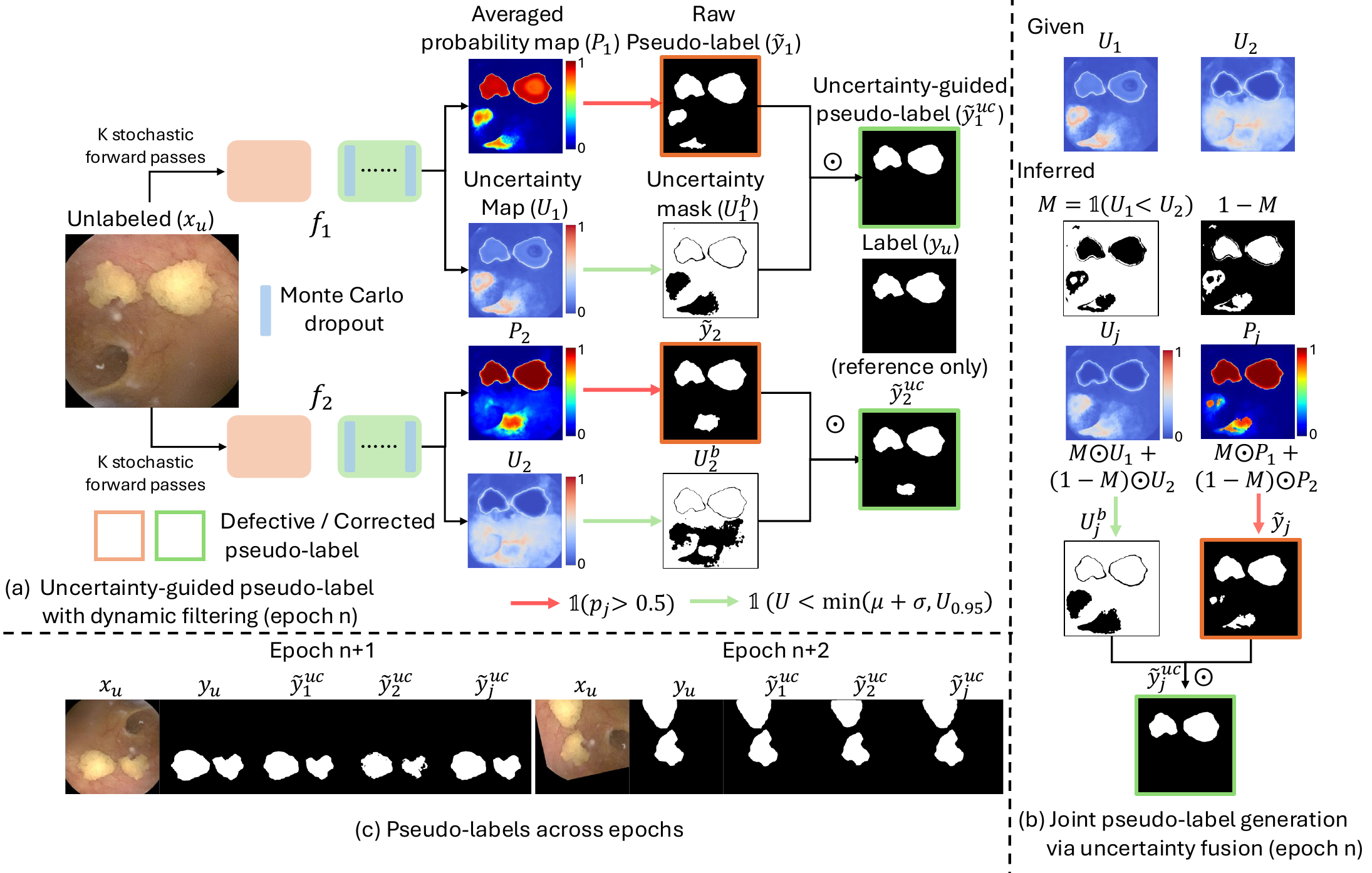}
\caption{
(a) For an unlabeled image $x_u$, uncertainty-guided pseudo-labels $\tilde{y}_1^{uc}$ and $\tilde{y}_2^{uc}$ (green boxes) are obtained by dynamically filtering the raw pseudo-labels $\tilde{y}_1$ and $\tilde{y}_2$, leading to cleaner supervision. The label $y_u$ of the unlabeled image is shown for reference only.
(b) $M$ chooses the lower-uncertainty prediction at each pixel to obtain the joint pseudo-label $\tilde{y}_j^{uc}$ for more reliable supervision by correcting residual defects in $\tilde{y}_2^{uc}$ from (a).
(c) Compared with the pseudo-labels at epoch $n$ in (a), the  $\tilde{y}_1^{uc}$, $\tilde{y}_2^{uc}$ and $\tilde{y}_j^{uc}$  at epochs $n+1$ and $n+2$ become cleaner and more consistent with $y_u$, indicating the effectiveness of (a) and (b).}

\label{uncertainty}
\end{figure}

\paragraph{Epistemic uncertainty.}
The cross-supervision setup naturally accommodates stochastic regularization, so we insert Monte Carlo dropout \cite{kendall2017uncertainties} layers after each decoder convolution to estimate uncertainty and improve the reliability of pseudo-labels, which further improves segmentation performance \cite{yu2019uncertainty}. Specifically, as shown in Fig.~\ref{uncertainty}(a), each unlabeled sample $x_u$ is passed through both networks multiple times to estimate entropy-based uncertainty. For each network $f_i$ $(i\in\{1,2\})$, the final output probability map is computed as $P_i=\frac{1}{K}\sum_{k=1}^{K} p_{i,k}$, where $p_{i,k}$ denotes the probability map in the $k$-th forward pass of network $i$, and we set $K=5$. The entropy-based epistemic uncertainty map is derived as $U_i=\frac{1}{K}\sum_{k=1}^{K} h(p_{i,k})$, with $h(p)=-p\log p-(1-p)\log(1-p)$.


\paragraph{Dynamic filtering.}Unlike previous works that use a fixed threshold \cite{sohn2020fixmatch}, the entire uncertainty map \cite{luo2022semi-uncertainty} or quantile-based selection \cite{yu2019uncertainty,yang2023revisiting}, we use a dynamic and data-driven thresholding strategy. Given $U_i$, the threshold is set as $T_i=\min[\mu(U_i)+\sigma(U_i),\, U_{i,0.95}]$, where $\mu$, $\sigma$ and $U_{i,0.95}$ denote the mean, standard deviation and $95^{th}$ percentile, respectively. Our adaptive thresholding approach effectively handles long-tail distributions and noisy predictions, yielding a more reliable uncertainty-based binary mask $U_i^{b}=\mathbbm{1}(U_i<T_i)$, where $\mathbbm{1}$ denotes the indicator function (see Fig.~\ref{uncertainty}(a)). The final uncertainty-guided pseudo-label for $x_u$ is then formulated as $\tilde{y}^{uc}_{i}=\tilde{y}_{i}\odot U_i^{b}$.

\subsection{Joint pseudo-label supervision} 

    Even with the incorporation of uncertainty estimates, the pseudo-labels may still be too noisy to provide appropriate supervision for harder samples. Most existing methods solely rely on the $\tilde{y}_u$ from each network for supervision, which may not be sufficient. To address this, \textit{our hypothesis is that joint supervision can effectively refine pseudo-labels by leveraging complementary information from both networks, providing more reliable supervision for challenging samples.}

As shown in Fig.~\ref{uncertainty}(b), the joint pseudo-label $\tilde{y}^{uc}_{j}$ is constructed in three steps: 
(1) Given the uncertainty maps $U_1$ and $U_2$ from the two networks in Endo-SemiS, we create a binary mask 
$M = \mathbbm{1}(U_1 < U_2)$ that selects the more confident prediction at each pixel. 
(2) Using this mask, we form the joint probability 
$P_j = M \odot P_1 + (1 - M) \odot P_2$ and obtain the raw pseudo-label $\tilde{y}_{j}$ by thresholding $P_j$ at $0.5$, while the joint uncertainty map is defined as 
$U_j = M \odot U_1 + (1 - M) \odot U_2$. 
 (3) Finally, we apply the dynamic filtering scheme to $U_j$ to obtain the binary uncertainty mask $U^b_j$ and compute the final uncertainty-guided joint pseudo-label as $\tilde{y}^{uc}_{j} = \tilde{y}_{j} \odot U^b_j$.

For an unlabeled image $x_u$ and its strongly augmented version $x_u^s$, 
we extend the cross-supervision loss in Eq.~\ref{pseudo_supervision_eq} to a weak–strong setting, 
where pseudo-labels are generated from the weak augmented image (see Sec.~\ref{AC}) and used to supervise the strongly augmented image. Together with uncertainty-guided pseudo-label learning,
the cross pseudo-supervised loss $L_p^{\text{cross}}(x_u, x_u^s)$ is defined as:
\begin{equation}
L_p^{\text{cross}}(x_u, x_u^s)
=
\underbrace{
    L_p\big(f_1(x_u), \tilde{y}^{uc}_{2}\big)
  + L_p\big(f_2(x_u), \tilde{y}^{uc}_{1}\big)
}_{\text{uncertainty-guided cross-supervision}}
+
\underbrace{
    L_p\big(f_1(x_u^s), \tilde{y}^{uc}_{j}\big)
  + L_p\big(f_2(x_u^s), \tilde{y}^{uc}_{j}\big)
}_{\text{joint pseudo-label supervision}}
\end{equation}

\subsection{Multi-level mutual learning}
Individual networks may independently learn different representations, which can cause divergence and inconsistencies in their predictions. If one network is consistently wrong, it can bias the other network and propagate errors. We propose a multi-level mutual learning approach to mitigate this variability by aligning the learning trajectories of both models and promoting consistency in their predictions. Although it does not guarantee correctness on unlabeled data, it reduces randomness and stabilizes the learning process, making models less likely to reinforce extreme errors.

We use the labeled data to apply mutual learning between the two networks. This encourages similarity at both the encoders and the decoders. The consistency from encoder and bottleneck features helps align feature representations and reduce variability in learned embeddings. 
Unlike previous work, which enforces the similarity between the probability maps \cite{zhang2018deep}, we enforce prediction consistency at the decoder level by aligning the logit maps of the networks, which is particularly important when generating pseudo-labels. Since pseudo-labels are filtered based on confidence thresholds, mutual learning stabilizes training by reducing prediction variance between networks, making the pseudo-label selection process more reliable.



For a labeled image $x_l$, let $f_1^e, f_1^b, f_1^l$ and $f_2^e, f_2^b, f_2^l$ denote the
first encoder feature maps, bottleneck features, and logit maps of the two networks, respectively.
The multi-level mutual learning loss is defined as:
\begin{equation}
L_m(x_l) =
 L_{\mathrm{ssim}}\!\big(f_1^e, f_2^e\big)
+  0.5\, \big( L_{\mathrm{kl}}(p_1^b \parallel p_2^b)
                  + L_{\mathrm{kl}}(p_2^b \parallel p_1^b) \big)
+ 2\, L_{\mathrm{mse}}\!\big(f_1^l, f_2^l\big)
\end{equation}
where $p_i^b = \mathrm{softmax}(f_i^b)$ denotes the channel-wise probability distribution of the bottleneck feature map, $i\in\{1,2\}$.

\paragraph{Total objective function.}
For labeled and unlabeled data, the total objectives are:
\begin{equation}
L(x_l) = L_s(x_l) + 0.5\, L_p^{\text{cross}}(x_l) + 0.5\, L_m(x_l), \quad L(x_u) = 0.5\, L_p^{\text{cross}}(x_u, x_u^s)
\end{equation}

\subsection{Spatiotemporal (ST) correction at frame level}
Segmentations produced on semi-supervised frames may exhibit frame-level inconsistencies due to the lack of temporal information, which appear as isolated false positive (FP) or false negative (FN) frames. As a post-processing step, we leverage the inherent spatiotemporal information in video clips, and introduce a separate correction model ($f_{st}$) at frame level to mitigate false positive FP and FN frames. 

We denote the $n^{th}$ test frame by $x_n$ and its predicted binary segmentation mask by $\tilde{y}_n$. For each frame $x_n$, we define $R_n$ as the total number of foreground pixels in $\tilde{y}_n$. Our key assumption is that adjacent frames should not exhibit large discrepancies in $R_n$. In particular, for FN frames, the target regions overlap across these frames, whereas for FP frames, the background region remains consistent (or contains little foreground). These assumptions motivate our inter-frame FP/FN detection and correction. We enforce temporal consistency by correcting FP frames when $R_n > 0$ and $R_{n-1} = R_{n+1} = 0$. Similarly, we classify $x_n$ as a FN frame when $R_n = 0$ and $R_{n-1} > r$ and $R_{n+1} > r$. We set $r = \tfrac{1}{4} H W$, where $H$ and $W$ denote the frame height and width.

To refine the predictions, we train a separate correction model $f_{st}$ that operates on a local temporal window.
Given labeled training pairs $\{(x_{n-2}, y_{n-2}), \ldots, (x_{n+2}, y_{n+2})\}$ sampled from $\{x_l, y_l\}$
, we concatenate them along the channel dimension to form $c_n$, and use this as input to predict a refined segmentation for the central frame $x_n$. During training, random corruptions are introduced to the masks with basic morphological operations or by setting them to zero. We use the MSE loss to enforce spatiotemporal consistency, and the total loss is: 
\begin{equation}
L = L_s(f_{st}(c_n), y_n)
  + 0.25 \sum_{k \in \{-1,1\}} L_{\mathrm{mse}}(f_{st}(c_n), y_{n+k})
  + 0.1  \sum_{k \in \{-2,2\}} L_{\mathrm{mse}}(f_{st}(c_n), y_{n+k})
\end{equation}
This formulation allows the network to leverage spatiotemporal information while preventing it from overly dominating the training process, thereby accommodating potential variations between frames. For inference, the correction model $f_{st}$ is applied to frames classified as FP or FN, and uses adjacent masks to satisfy the local-consistency assumption for challenging ureteroscopy videos.

\section{Experiments}


\paragraph{{Kidney stone dataset.}} This in-house dataset \cite{deol2024mp07} consists of 38 fiberoptic and 98 digital endoscopy videos. We extracted frames at 3 FPS, resulting in a total of 21,718 labeled frames. We partitioned the data at the video-level, yielding approximately a 75/5/20\% split for training/validation/testing. While all videos contain kidney stones,  some individual frames may not. This adds complexity to the segmentation task, as it also introduces an implicit detection challenge. The dataset exhibits large variation in image quality due to the complex in vivo environment during surgery (Fig.~\ref{dataset}). The images are resized into $256\times 256$.

\paragraph{{Polyp colonoscopy dataset.}} PolypGen \cite{ali2023multi} is a publicly available multi-center dataset with 1,537 single-labeled frames (discrete sampling) and 2,225 sequence-labeled frames (short clips) collected from six different imaging centers. Following the benchmark study \cite{ali2023multi}, we use data from centers 1–5 for training and test on center 6. We resize images to $512\times 512$.

\paragraph{Implementation details.} During training, we set the $L_s$ and $L_p$ as naive binary cross entropy loss with a batch size of 16 for 200 epochs. The initial learning rate is $10^{-4}$ with a cosine curve decay to $10^{-5}$. Our study was conducted on an NVIDIA A6000.

\paragraph{Compared methods.}
We compare to several state-of-the-art semi-supervised segmentation methods, including 
Generic \cite{bellver2019budget}, AllSpark \cite{wang2024allspark},
UPRC \cite{luo2022semi-uncertainty},
FixMatch \cite{sohn2020fixmatch}, UniMatch \cite{yang2023revisiting}, Mean Teacher \cite{tarvainen2017mean},
Cross-Pseudo Supervision (CPS) \cite{chen2021semi} and Cross Teaching \cite{luo2022semi-crossteaching}. 

These methods can be categorized into single-network (Generic, AllSpark, UPRC, FixMatch, UniMatch, Mean Teacher) and cross-supervision (CPS and Cross Teaching) methods, and some of these approaches  incorporate transformer-based architectures, such as Cross Teaching, AllSpark. These methods explore different forms of uncertainty modeling, including aleatoric uncertainty (AllSpark, FixMatch, UniMatch) and epistemic uncertainty (UniMatch, MeanTeacher, UPRC). Most approaches rely on pseudo-labeling (FixMatch, UniMatch, CPS, CrossTeaching, AllSpark) and uncertainty-guided self-consistency mechanisms (MeanTeacher, UPRC) to improve learning stability and reliability. We implemented these methods with their official code repositories. Further details on the category classification of the compared methods are provided in Appendix~\ref{appendix:compared methods}.

\paragraph{Evaluation metrics.}
We report pixel-level segmentation performance using Dice, sensitivity, and specificity. We also evaluate image-level target presence detection by converting each predicted mask into a binary image label. An image is predicted positive if any foreground pixel is present and negative otherwise. The precision, recall, F1-score, and accuracy are computed at the image level. These metrics indicate whether the model detects the presence or absence of the target object, independent of pixel-wise overlap quality. 




\begin{table*}[t]
\small
\caption{Kidney results ($mean\pm stdev.$, in \%)  with \textbf{10\% labeled data}. Bold indicates the \textbf{best}. The horizontal sections show: supervised (gray), semi-supervised with single network (blue), cross-supervised (lavender),  and supervised with 100\% labeled data, i.e., upper bound (green). Our method achieved the highest Dice score, sensitivity, F1, and accuracy.} 

\label{main_table}
\begin{center}
    \begin{tabular}{ l  | c |c |c |c |c |c |c}
    \toprule

    \multicolumn{1}{c}{ } & \multicolumn{3}{c|}{Pixel-level} & \multicolumn{4}{c}{Image-level} \\ 
    \hline
    \multicolumn{1}{l}{Methods} &  \multicolumn{1}{c}{Dice} & \multicolumn{1}{c}{Sensitivity} & \multicolumn{1}{c|}{Specificity} &  \multicolumn{1}{c}{Pre.} & \multicolumn{1}{c}{Rec.} &  \multicolumn{1}{c}{F1} & \multicolumn{1}{c}{Acc.}\\

            
    \hline
    \rowcolor{supervised}
    U-Net  & 80.5$\pm$32.1 
    & 88.6$\pm$22.0  & 95.4$\pm$8.4
            & 88.7 & 95.3 & 92.8 & 90.1 \\
    \rowcolor{supervised}
    nnU-Net  & 79.5$\pm$33.8 & 85.9$\pm$27.4 & 95.5$\pm$9.1
            & 90.1 &91.1  & 90.6 & 87.6\\
    \hline
    \rowcolor{semisup}
    Generic  & 78.5$\pm$31.7 & 86.1$\pm$25.7 & 92.3$\pm$13.9
            & 90.7 & 95.3 & 92.9 & 90.5 \\
    \rowcolor{semisup}
    AllSpark  & 77.0$\pm$31.2 & 88.0$\pm$24.8 & 89.3$\pm$18.0
            & 94.7 & 92.8 
            & 93.8 & 91.7 \\
    \rowcolor{semisup}
    UPRC  & 80.7$\pm$31.4 
    & 84.0$\pm$27.3  & 96.4$\pm$7.8
    & 92.9 &  94.6
            &  93.7 & 91.6 \\
  \rowcolor{semisup}  
    FixMatch  & 81.9$\pm$31.7 
    & 89.8$\pm$22.4 & 94.3$\pm$10.9
    & 89.7 & \textbf{96.5}  
            & 93.0 & 90.5 \\
  \rowcolor{semisup}  
    UniMatch  & 85.5$\pm$27.6 
    & 89.4$\pm$23.2 & 95.5$\pm$8.9
    & 94.3 &  96.4
            & 95.4 & 91.7 \\
    \rowcolor{semisup}
    Mean Teacher & 82.2$\pm$31.2  & 84.1$\pm$28.6 & 96.6$\pm$8.5
    & 95.6 &  90.5
            & 93.0 & 91.1 \\
    \hline    
    \rowcolor{crosssup}
    CPS & 85.2$\pm$28.0 & 88.8$\pm$22.8 & 95.8$\pm$8.8
            & 94.0 & 96.1
            & 95.0 & 93.4 \\
    \rowcolor{crosssup}
    Cross Teaching & 85.6$\pm$28.7 & 87.6$\pm$26.5 & \textbf{96.7$\pm$7.4}
            & \textbf{96.5} & 92.6 
            & 94.8 & 92.9 \\
        
\rowcolor{crosssup}
    Endo-SemiS (Ours) & \textbf{87.6$\pm$26.4} & \textbf{91.1$\pm$21.5} & 96.0$\pm$8.4
            &  95.0 &  96.1 
            &  \textbf{95.6} & \textbf{94.1} \\
        
    \hline
    \rowcolor{upper}
    Upper bound U-Net  & 85.3$\pm$29.2 & 89.0$\pm$24.5 & 96.5$\pm$8.2
            & 94.4 & 94.2 & 94.3 & 92.5\\
    \rowcolor{upper}
    Upper bound nnU-Net  & 85.5$\pm$28.5 & 89.3$\pm$24.5 & 96.0$\pm$8.6
            & 92.4 & 93.3 & 92.9 & 90.5\\

    \bottomrule

\end{tabular} 

\end{center}
\end{table*}

\begin{figure}[t]
\centering
\includegraphics[width=0.9\linewidth]{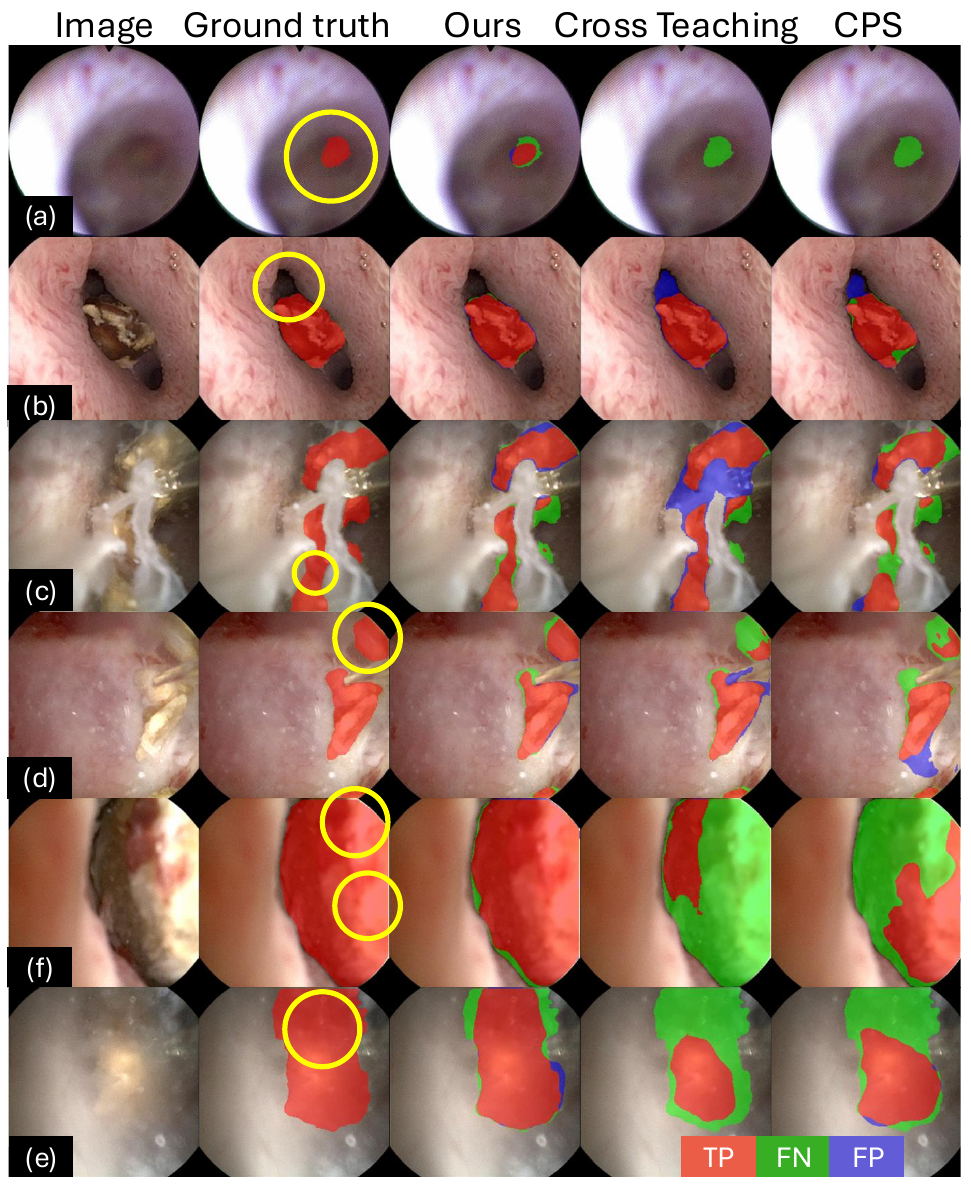}
\caption{Qualitative kidney stone results (10\% labeled data). Yellow circles highlight poor visibility areas. (a) fiberoptic frames, (b) digital frames, (c) fluid distortions,  (d) motion blur, (e) debris during stone ablation, and (f) illumination changes. 
}
\label{qualitative}
\end{figure}

\paragraph{Segmentation performance.} The quantitative results of the kidney stone dataset using 10\% labeled data are shown in Tab.~\ref{main_table}. The Generic model underperforms compared to supervised learning, which highlights the critical role of pseudo-label quality in semi-supervised segmentation.
In contrast, the results of Mean Teacher, UniMatch, and FixMatch show that incorporating external uncertainty  improves segmentation, especially for UniMatch where epistemic uncertainty is also leveraged. The results of AllSpark indicate that transformer-based method struggles for kidney stone segmentation, where image quality is variable (Fig.~\ref{qualitative}). Cross-supervision methods (lavender) achieve better performance than single-network-based methods (blue), demonstrating better generalizability. 
Endo-SemiS achieves substantially superior performance across most metrics compared to these SOTA semi-supervised  methods. Notably, it even outperforms supervised methods trained on full labeled data (upper bound, green).

\paragraph{Consistency analysis.} In Tab.~\ref{ratio_table}, we present consistency results in two aspects: (1) robustness across different ratios of labeled training data, and (2) consistency between models within the framework. Endo-SemiS maintains stable performance across different ratios, demonstrating particularly robust performance when labeled data is extremely limited (only 1\%). The performance of the two cross-supervised models of our framework is more consistent and reliable than the compared methods.
Considering the challenging visibility conditions in kidney stone surgery (Fig.~\ref{qualitative}), consistency is crucial to performance because inaccurate  pseudo-labels can severely degrade segmentation results.
Finally, we observe that our ST corrective model improves performance across all label ratios.

\begin{table*}[t]
\small
\caption{Dice (\%) on kidney dataset with various labeled data ratios. ``-1'' and ``-2'' denote individual networks for cross-supervision. ST: spatiotemporal correction. Bold indicates the best in each category. Lavender denotes the cross-supervised methods.} 

\label{ratio_table}
\begin{center}
    \begin{tabular}{ l  | c |c |c |c | c}
    \toprule


    \multicolumn{1}{l}{Methods} &  \multicolumn{1}{c}{$1\%$} & \multicolumn{1}{c}{$5\%$}  &  \multicolumn{1}{c}{$10\%$} & \multicolumn{1}{c}{$30\%$} & \multicolumn{1}{c}{$100\%$}\\

    \hline
    U-Net 
          & 74.9$\pm$34.1 
          & 77.8$\pm$34.5 
          & 80.5$\pm$32.1
          & 82.0$\pm$32.0 
          & 85.3$\pm$29.2\\

    nnU-Net  & 76.4$\pm$34.3 & 78.0$\pm$34.5 & 79.5$\pm$33.8 &  
         82.1$\pm$31.6 & 85.5$\pm$28.5 \\
    \hline
    Generic  & 69.4$\pm$37.3 & 76.5$\pm$34.3 & 78.5$\pm$31.7 & 83.4$\pm$29.6 &- \\
    \hline
        \rowcolor{crosssup}

    CPS-1  & 82.9$\pm$30.5 & 84.7$\pm$28.8 & 85.2$\pm$28.0 & 85.7$\pm$27.7 &- \\
    \rowcolor{crosssup}

    Cross Teaching-1 & 77.1$\pm$32.4 & 80.1$\pm$32.2 & 85.6$\pm$28.7 & 86.5$\pm$27.6 & -\\
        \rowcolor{crosssup}

    Endo-SemiS-1 & \textbf{86.5$\pm$27.6} & \textbf{87.5$\pm$26.4} & \textbf{87.6$\pm$26.4} &  \textbf{87.9$\pm$26.1} & - \\

    \hline
    CPS-1+ST & 83.8$\pm$29.5 & 85.3$\pm$28.1 &  85.7$\pm$27.4 &  86.2$\pm$27.1
            & - \\
          
    Endo-SemiS-1+ST & \textbf{87.1$\pm$27.1} & \textbf{87.8$\pm$26.3} &  \textbf{88.1$\pm$25.7} & \textbf{88.2$\pm$25.8} 
            & - \\

    \bottomrule
    \multicolumn{6}{c}{Performance variability in cross-supervised segmentation ($\pm$Dice in \%)} \\
    \hline
    CPS-2  & -1.0 & +1.9 & -0.7 & +0.6 &- \\

    Cross Teaching-2  & -11.4 & -13.6 & -4.0 & -4.6 & -\\
    
    Endo-SemiS-2 & \textbf{-0.9} & \textbf{+0.1} & \textbf{0.0} & \textbf{-0.2}  & - \\
    \bottomrule

\end{tabular} 

\end{center}
\end{table*}

\begin{table*}[t]
\small
\caption{Dice (\%) for ablation study on kidney dataset with 10\% labeled data. AU, EU: aleatoric/epistemic uncertainty. JPS: joint pseudo-label supervision. ML-D: mutual learning in decoder. ML-EB: mutual learning in encoder and bottleneck. Bold denotes the best for each model.} 

\label{ablation_table}
\begin{center}
    \begin{tabular}{ l  | c |c |c |c | c |c }    
    \toprule

 
    & baseline \cite{chen2021semi}
    & + AU
    & + EU
    & + JPS
    & + ML-D
    & + ML-EB
    \\
    
    \hline
    Endo-SemiS-1 
    & 85.2 
    & 86.2 
    & 86.9 
    & \textbf{87.8} 
    & 87.2 
    & 87.6 
    \\
    \hline
    Endo-SemiS-2 
    & 84.5 
    & 86.4 
    & 87.2 
    & 86.8 
    & 87.5 
    & \textbf{87.6} 
    \\
    
    \bottomrule

\end{tabular} 

\end{center}
\end{table*}

\paragraph{Ablation analysis.} Tab.~\ref{ablation_table} shows the ablation study, where CPS is used as the baseline method, and the improvements for each added component are shown. Importantly, joint pseudo-label supervision (JPS) yields a larger improvement, which indicates that it effectively removes uncertain regions and generates high-quality pseudo-labels for supervision, especially for strong augmented images.
Although multi-level mutual learning slightly decreases the performance, it improves consistency.

\begin{table*}[t]
\small
\caption{Quantitative results (\%) in polyp dataset with 10\% labeled data. The left and right parts show the results for single and sequence frames, respectively. } 

\label{polyp_table}
\begin{center}
    \begin{tabular}{ l  | c |c |c |c |c |c }
    \toprule

    \multicolumn{1}{l}{} &  \multicolumn{3}{c}{Single frame data} & \multicolumn{3}{c}{Sequence frame data} \\
    \hline
    \multicolumn{1}{l}{Methods} &  \multicolumn{1}{c}{Dice} & \multicolumn{1}{c}{Sensitivity} & \multicolumn{1}{c|}{Specificity} &  \multicolumn{1}{c}{Dice} & \multicolumn{1}{c}{Sensitivity} & \multicolumn{1}{c}{Specificity}\\

    \hline
    U-Net & 75$\pm$\textbf{30} & 73$\pm$31 & \textbf{100$\pm$1}
            & 64$\pm$38 & 64$\pm$38 & \textbf{100$\pm$1}  \\
    \hline
            
    Endo-SemiS-1 & 76$\pm$34 & 75$\pm$34 & \textbf{100$\pm$1}
    & 69$\pm$39 &  67$\pm$39
            &   \textbf{100$\pm$1} \\

    Endo-SemiS-2 & \textbf{79$\pm$30} & 77$\pm$31 & \textbf{100$\pm$1}
    & \textbf{71$\pm$37} &  70$\pm$37
            &  \textbf{100}$\pm$2  \\

    \hline
    Upper bound U-Net & \textbf{79$\pm$30}* & \textbf{79$\pm$31} & 99$\pm$2
            & 69$\pm$\textbf{37}* & \textbf{74$\pm$35} & 99$\pm$2  \\

    \bottomrule
\multicolumn{7}{l}{
\begin{tabular}{@{}l@{}@{}} 
    * denotes our implementation; benchmark \cite{ali2023multi} results are 79\% and 66\%.
\end{tabular}
    } 


\end{tabular} 

\end{center}
\end{table*}

\paragraph{Generalizability analysis.} We also evaluate the proposed Endo-SemiS on the polyp segmentation task (Tab.~\ref{polyp_table}).
The results using only 10\% labeled data show that Endo-SemiS  outperforms supervised methods (U-Net) and reaches the upper bound (single frame data) as well as surpasses it (sequence frame data). Furthermore, the  performance of the two models is consistent, showing the robustness of our approach to the domain shift between different imaging sites.


\section{Conclusion}
In this study, we propose \textbf{Endo-SemiS} for robust endoscopic segmentation via semi-supervised learning under limited annotation.   Endo-SemiS extends cross-supervision by integrating uncertainty-guided pseudo-label generation, joint pseudo-label supervision, and multi-level mutual learning to improve training stability and pseudo-label reliability.  We evaluate Endo-SemiS on two clinical endoscopy applications, kidney stone laser lithotomy from ureteroscopy and polyp screening from colonoscopy, using two datasets with challenging image quality.  Compared to state-of-the-art semi-supervised segmentation methods, Endo-SemiS achieves superior segmentation performance, indicating improved robustness and generalization under challenging endoscopic conditions. In addition, a spatiotemporal corrective network further improves performance by leveraging inter-frame information.  Future work will apply Endo-SemiS to additional endoscopic procedures and broader domain shifts, and will further incorporate temporal information into the semi-supervised learning framework.

\clearpage  
\midlacknowledgments{This work was supported in part by the National Institutes of Health (R21DK133742) and Vanderbilt Institute for Surgery
and Engineering (VISE) Seed Grant. Daiwei Lu is supported by NIH F31DK143735-01.}

\bibliography{midl-samplebibliography}
\appendix

\section{Compared methods}
\label{appendix:compared methods}
We compare our method with several state-of-the-art semi-supervised approaches \cite{wang2024allspark,sohn2020fixmatch,yang2023revisiting,tarvainen2017mean,luo2022semi-uncertainty,chen2021semi,luo2022semi-crossteaching}. 
These methods cover both single-network \cite{wang2024allspark,sohn2020fixmatch,yang2023revisiting,tarvainen2017mean} and cross-supervision \cite{chen2021semi,luo2022semi-crossteaching} frameworks, with and without transformer backbones \cite{luo2022semi-crossteaching,wang2024allspark}. 
They focus on different uncertainty modeling strategies, including aleatoric \cite{wang2024allspark,sohn2020fixmatch,yang2023revisiting} and epistemic \cite{yang2023revisiting,tarvainen2017mean,luo2022semi-uncertainty} uncertainty, and combine confidence-based pseudo-labeling \cite{sohn2020fixmatch,yang2023revisiting,chen2021semi,luo2022semi-crossteaching,wang2024allspark} with uncertainty-guided self-consistency \cite{tarvainen2017mean,luo2022semi-uncertainty}. 
For completeness, we summarize the main characteristics of each method below.

\begin{itemize}

\item \textbf{AllSpark} \cite{wang2024allspark}:
Single-network transformer-based semi-supervised semantic segmentation method built on a standard pseudo-labeling baseline. It inserts an AllSpark bottleneck between the encoder and decoder, where channel-wise cross-attention and a class-wise semantic memory reconstruct labeled features from unlabeled features to strengthen supervision.
It was published at \textit{CVPR} 2024.

\item \textbf{Uncertainty-Rectified Pyramid Consistency (URPC)} \cite{luo2022semi-uncertainty}:
It is a single-network pyramid-prediction framework for semi-supervised medical image segmentation.
The model produces multi-scale predictions and, for unlabeled data, enforces consistency between each scale and their average prediction.
Pixel-wise uncertainty is estimated from the discrepancy among scales in a single forward pass and is used both to weight the pyramid consistency loss and to impose an uncertainty-minimization regularizer, enabling more reliable use of unlabeled images. It was published at \textit{Medical Image Analysis} 2022.

\item \textbf{FixMatch} \cite{sohn2020fixmatch}:
Single-network method with a CNN backbone that combines consistency regularization and pseudo-labeling. 
For each unlabeled image, it takes the prediction on a weakly augmented view, keeps it as a qulified pseudo-label only if its confidence exceeds a fixed threshold, and trains the model to match this pseudo-label on a strongly augmented view of the same image. It was published at \textit{NeurIPS} 2020.

\item \textbf{UniMatch} \cite{yang2023revisiting}:
Single-network method with a CNN backbone that revisits FixMatch for semi-supervised semantic segmentation.
It maintains weak-strong consistency using fixed confidence-thresholded pseudo-labels from the weakly augmented image, and introduces unified perturbations that operate at both the image level (strong augmentations) and the feature level (dropout), together with two strongly augmented images guided by the same weak prediction, to better exploit the perturbation space. It was published at \textit{CVPR} 2023.

\item \textbf{Mean Teacher} \cite{tarvainen2017mean}: 
Teacher-Student framework with a single concolutional neural network (CNN) backbone. The student is trained on labeled data, and an exponential moving average (EMA) of the student weights defines the teacher. For unlabeled data, a consistency loss enforces that the student prediction matches the teacher prediction under stochastic perturbations. This can be viewed as reducing epistemic uncertainty. It was published at \textit{NeurIPS} 2017.

\item \textbf{Cross Pseudo Supervision (CPS)} \cite{chen2021semi}:
Cross-supervision semi-supervised semantic segmentation framework in which two segmentation networks with the same architecture but different initializations are trained jointly. For both labeled and unlabeled images, the prediction from each network is used as a pseudo label to supervise the other, enforcing prediction consistency and effectively expanding the training data.
It was published at \textit{CVPR} 2021.

\item \textbf{Cross Teaching between CNN and Transformer (Cross Teaching)} \cite{luo2022semi-crossteaching}:
Cross-supervision semi-supervised segmentation framework that pairs a CNN (UNet) and a Transformer (Swin-UNet).
On unlabeled images, each network takes the prediction from the other network as a pseudo-label and is optimized with a cross-teaching Dice loss, providing implicit consistency while exploiting the complementary local and long-range representations of CNNs and transformers.
It was published at \textit{MIDL} 2022.

\end{itemize}

\end{document}